\documentclass[10pt,twocolumn,letterpaper]{article}

\usepackage{iccv}
\usepackage{times}
\usepackage{epsfig}
\usepackage{graphicx}
\usepackage{amsmath}
\usepackage{amssymb}
\usepackage{algorithm}
\usepackage{algorithmic}
\usepackage{overpic}
\usepackage{soul}
\usepackage{caption}
\usepackage{subcaption}
\usepackage{booktabs}
\usepackage[toc,page]{appendix}
\usepackage[numbers,sort,compress]{natbib}
\usepackage[pagebackref=true,breaklinks=true,letterpaper=true,colorlinks,bookmarks=false]{hyperref}

\usepackage[capitalize]{cleveref}
\iccvfinalcopy % *** Uncomment this line for the final submission

\def\iccvPaperID{****} % *** Enter the ICCV Paper ID here

\ificcvfinal\fi

\newcommand{\methodname}{\mbox{Pix2Video}\xspace}
\newcommand{\totaluser}{37\xspace}
\newcommand{\eachuser}{11\xspace}

\begin{document}

%%%%%%%%% TITLE
\title{Pix2Video: Video Editing using Image Diffusion}

\author{Duygu Ceylan\textsuperscript{1*}\quad 
        Chun-Hao P. Huang\textsuperscript{1*}\quad 
        Niloy J. Mitra\textsuperscript{1,2}\\ 
\textsuperscript{1}Adobe Research \quad \textsuperscript{2}University College London \\ 
{\normalsize \url{https://duyguceylan.github.io/pix2video.github.io/}}%
%\vspace{-.25cm}
}

\maketitle
\ificcvfinal\thispagestyle{empty}\fi
\def\thefootnote{*}\footnotetext{These authors contributed equally to this work}

\begin{abstract}

Image diffusion models, trained on massive image collections,  have emerged as the most versatile image generator model in terms of quality and diversity. They support inverting real images and conditional (e.g., text) generation, making them attractive for high-quality image editing applications. We investigate how to use such pre-trained image models for text-guided video editing. The critical challenge is to achieve the target edits while still preserving the content of the source video. Our method works in two simple steps: first, we use a pre-trained structure-guided (e.g., depth) image diffusion model to perform text-guided edits on an anchor frame; then, in the key step, we progressively propagate the changes to the future frames via self-attention feature injection to adapt the core denoising step of the diffusion model. We then consolidate the changes by adjusting the latent code for the frame before continuing the process. Our approach is training-free and generalizes to a wide range of edits. We demonstrate the effectiveness of the approach by extensive experimentation and compare it against four different prior and parallel efforts (on ArXiv). 
We demonstrate that realistic text-guided video edits are possible, without any compute-intensive preprocessing or video-specific finetuning. 

\end{abstract}


\if0
\begin{teaserfigure}
  %\includegraphics[width=\textwidth]{figures/teaser}
  \vspace{1.5in}
  \caption{Todo.}
  \label{fig:teaser}
\end{teaserfigure}
\fi


\section{Introduction}

\begin{figure}[t!]
    \centering
    \includegraphics[width=\columnwidth]{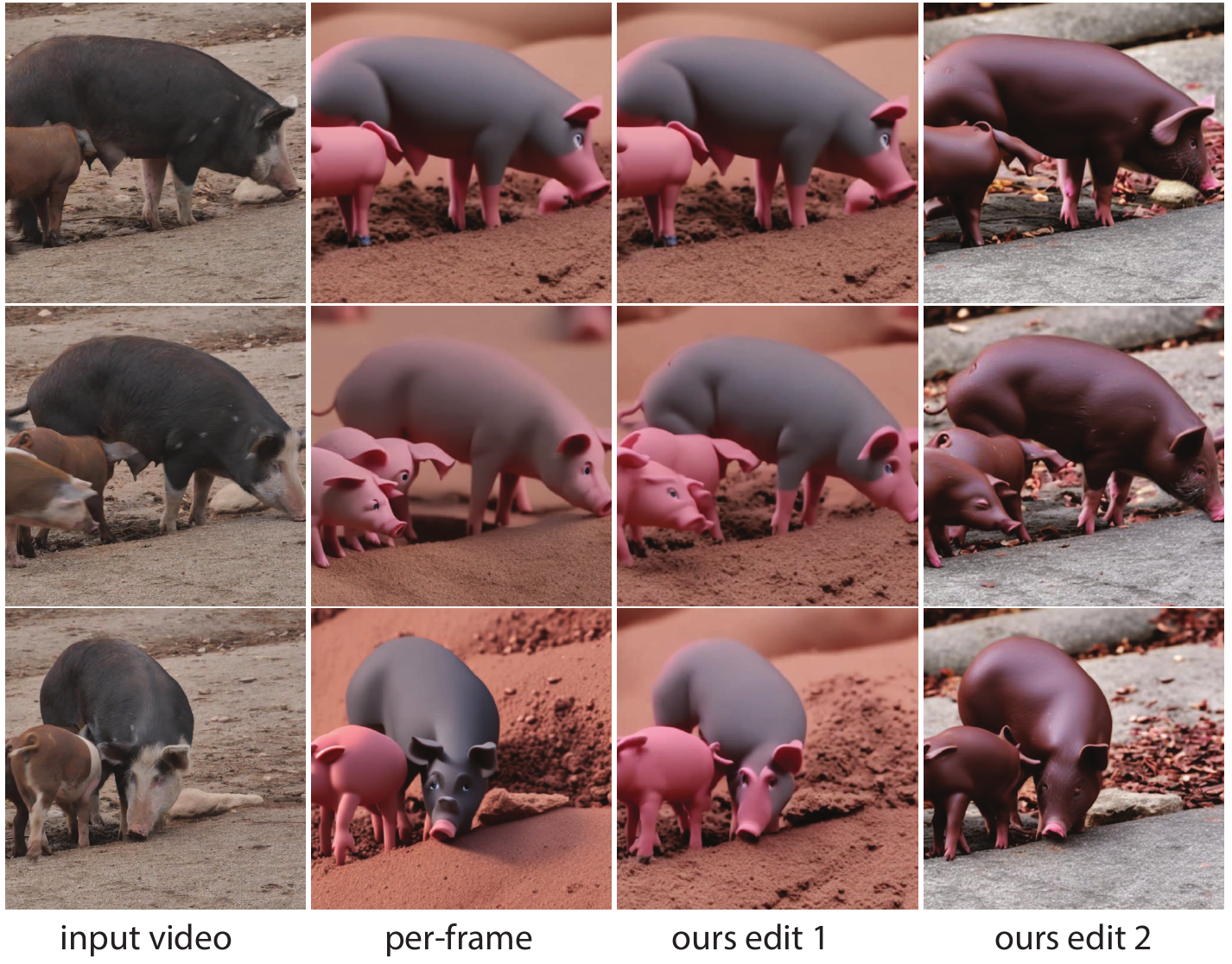}
    \caption{There has been exciting advancements in large scale image generation models~\cite{rombach2022stablediffusion}. When applied independently to a sequence of images (`per-frame'), however, such methods produce inconsistent results across frames. Our method uses a pre-trained and fixed image generation model to consistently edit a video clip based on a target text prompt. We show examples of two different edits (`ours').}
    \label{fig:motivation}
\end{figure}

% image-based diffusion + edits
Diffusion-based algorithms~\cite{croitoru2022diffusion,ho2020ddpm,song2021ddim} have emerged as the  generative model of choice for image creation.  They 
are stable to train (even over huge image collections), 
produce high-quality results, and support conditional sampling. Additionally, one can invert~\cite{mokady2022null,song2021ddim} a given image into a pretrained diffusion model and subsequently edit using only textual guidance~\cite{hertz2022prompt}. Such a generic workflow, to handle real images and interact using semantic text prompts, is an exciting development and opens the door for many downstream content creation tasks.

% nothing available for videos
However, the same workflow is barely available for videos where the development of video diffusion models is still in its infancy~\cite{Dreamix,singer2023makeavideo,yu2023video}. Not surprisingly,  naively applying an image-based workflow to each video frame  produces inconsistent results (see Figure~\ref{fig:motivation}). Alternately, while it is possible to use a single frame for style guidance and employ video stylization propagation~\cite{Jamriska19-SIG}, the challenge lies in stylizing new content revealed under changing occlusions across  frames.

% what are the challenges
In this paper, we explore the feasibility of \textit{editing a video clip using a pre-trained image diffusion model and text instructions with no additional training}. We start by inverting the input video clip and expecting the user to edit, using textual prompts, one of the video frames. The goal is then to \textit{consistently} propagate the edit across the rest of the video. The challenge is to balance between respecting the user edit and maintaining the plausibility and temporal coherency of the output video. Image generation models already generate images faithful to the edit prompt. %The alignment to the input text can further be improved as shown by recent methods~\cite{hertz2022prompt,pnpDiffusion2022}.
Hence, what remains challenging is to propagate the edit in a temporally coherent manner. 

Temporal coherency requires preserving the appearance across neighboring frames while respecting the motion dynamics. Leveraging the fact that the spatial features of the self attention layers are influential in determining both the structure and the appearance of the generated images, we propose to inject features obtained from the previously edited frames into the self attention layer of the current frame. This feature injection notably adapts the self attention layers to perform cross-frame attention and enables the generation of images with coherent appearance characteristics. To further improve consistency, we adopt a \textit{guided diffusion} strategy in which we update the intermediate latent codes to enforce similarity to the previous frame before we continue the diffusion process. While the image generation model cannot reason about motion dynamics explicitly, recent work has shown that generation can be conditioned on static structural cues such as depth or segmentation maps~\cite{zhang2023controlnet}. Being disentangled from the appearance, such structural cues provide a path to reason about the motion dynamics. Hence, we utilize a depth-conditioned image generation model and use the predicted depth from each frame as additional input.

We term our method \emph{\methodname} and evaluate it on various real video clips demonstrating both local (e.g., changing the attribute of a foreground object) and global (e.g., changing the style) edits. We compare with several state-of-the-art approaches including diffusion-based image editing methods applied per frame~\cite{meng2021sdedit}, patch-based video based stylization methods~\cite{Jamriska19-SIG}, neural layered representations that facilitate consistent video editing~\cite{bar2022text2live}, and concurrent diffusion based video editing methods~\cite{wu2022tuneavideo}. We show that \methodname is on par with or better than the baselines \textit{without} requiring any compute-intensive preprocessing~\cite{bar2022text2live} or any video-specific finetuning~\cite{wu2022tuneavideo}. 
This can be seen in Figure \ref{fig:motivation} ``ours'' columns where the appearance of the foreground objects are more consistent across frames than per-frame editing. 
% \duygu{discuss teaser}

In summary, we present a \textit{training free} approach that utilizes pre-trained large scale image generation models for video editing. Our method does not require pre-processing and does not incur any additional overhead during inference stage. This ability to use an existing image generation model paves the way to bring exciting advancements in controlled image editing to videos at no cost.

%just changing the conditioning signal is not enough, either early in the diffusion process (too many changes) or later in the diffusion process (view/time not taken into consideration); talk about null embedding

% main hypothesis

\section{Related Work}
\subsection{Image generation and editing}
While many deep generative models, e.g., GAN~\cite{NIPS2014_5ca3e9b1}, have demonstrated the ability to generate realistic images \cite{brock2018large,stylegan}, recently, diffusion models have become the choice of models due to the high quality output they achieve on large scale datasets~\cite{NEURIPS2021_49ad23d1}. 
Denoising Diffusion Probabilistic Model (DDPM) \cite{ho2020ddpm} and its variant Denoising Diffusion Implicit Model (DDIM) \cite{song2021ddim} have been  widely used for unconditional text-to-image generation. Several large scale text-to-image generation models~\cite{nichol2022glide,ramesh2022dalle2,saharia2022imagen}, which operate on the pixel space have been presented,  achieving very high quality results. Rombach et al.~\cite{rombach2022stablediffusion} have proposed to work in a latent space which has lead to the widely adopted open source Stable Diffusion model. We refer the readers to a recent survey~\cite{croitoru2022diffusion} and the extensive study~\cite{Karras2022edm} for a detailed discussion on diffusion models.

In the presence of high quality text conditioned image generation models, several recent works have focused on utilizing additional control signals for generation or editing existing images. Palette~\cite{saharia2022palette} has shown various image-to-image translation applications using a diffusion model including colorization, inpainting, and uncropping. Several methods have focused on providing additional control signals such as sketches, segmentation maps, lines, or depth maps by adapting a pretrained image generation model. These methods work by either finetuning~\cite{wang2022pretraining} an existing model, introducing adapter layers~\cite{t2iadapter} or other trainable modules~\cite{voynov2022sketch,zhang2023controlnet}, or utilizing an ensemble of denoising networks~\cite{ediffi}. Since our model uses a pretrained image diffusion model, it can potentially use any model that accepts such additional control signals. In another line of work, methods have focused on editing images while preserving structures via attention layer manipulation~\cite{hertz2022prompt,pnpDiffusion2022}, additional guidance optimization~\cite{Choi2021ILVR}, or per-instance finetuning~\cite{kawar2022imagic}. Our method also performs attention layer manipulation, specifically in the self attention layers, along with a latent update at each diffusion step. Unlike single image based editing work, however, we utilize previous frames when performing these steps. We would also like to emphasize that the edit of the anchor frame for our method can potentially be performed with any such method that utilize the same underlying image generation model.
%GLIDE \cite{nichol2022glide} propose cfg. Talk about DALLE-2 \cite{ramesh2022dalle2} and ImaGen \cite{saharia2022imagen}. Stable Diffusion \cite{rombach2022stablediffusion} improves stability by moving from pixel to latent space.
%SDEdit \cite{meng2021sdedit}, ILVR \cite{Choi2021ILVR}, Imagic \cite{kawar2022imagic},
%Prompt-to-prompt \cite{hertz2022prompt}, null inversion \cite{mokady2022null}, Plug-and-play \cite{pnpDiffusion2022},
%ControlNet \cite{zhang2023controlnet}

\subsection{Video generation and editing}
Until recently, GANs have been the method of choice for video generation, with many works designed towards unconditional generation~\cite{brooks2022generating,Gupta_2022_CVPR,TGAN2017,stylegan-v,MoCoGAN,digan,zhang2022towards}. In terms of conditional generation, several methods have utilized guidance channels such as segmentation masks or keypoints~\cite{wang2018fewshotvid2vid,wang2018vid2vid}. However, most of these methods are trained on specific domains. One particular domain where very powerful image generators such as StyleGAN~\cite{stylegan} exist is faces. Hence, several works have explored generating videos by exploring the latent space of such an image based generator~\cite{StyleFaceV,xu2022videoeditgan}. While we also exploit an image generation model, we are not focused on a specific domain.

With the success of text-to-image generation models, there has been recent attempts in text-to-video generation models using architectures such as transformers~\cite{hong2022cogvideo,villegas2023phenaki,MAGVIT} or diffusion models~\cite{he2022lvdm,imagen-video,singer2023makeavideo,yu2023video}. However, such models are still in their infancy compared to images, both due to the complexity of temporal generation as well as large scale annotated video datasets not being comparable in size to images. Concurrent works~\cite{esser2023structure,Dreamix} explore mixed image and video based training to address this limitation. In another concurrent work, Wu et al.~\cite{wu2022tuneavideo} inflate an image diffusion model and finetune on a specific input video to enable editing tasks. In our work, we use the pretrained image diffusion model as it is with no additional training.

Layered neural representations~\cite{lu2020} have recently been introduced, providing another direction for editing videos. Layered neural atlases~\cite{kasten2021layered} are such representations that map the foreground and background of a video to a canonical space. Text2Live~\cite{bar2022text2live} combines such a representation with text guidance to show compelling video editing results. While impressive, the computation of such neural representations includes extensive per-video training (7-10 hours),  which limits their applicability in practice.

Finally, video stylization is a specific type of editing task where the style of an example frame is propagated to the video. While some methods utilize neural feature representations to perform this task~\cite{RuderDB2016}, Jamriska et al.~\cite{Jamriska19-SIG} consider a patch-based synthesis approach using optical flow. In a follow-up work~\cite{Texler2020}, they provide a per-video fast patch-based training setup to replace traditional optical flow. Both methods achieve high quality results but are limited when the input video shows regions that are not visible in the provided style keyframes. They rely on having access to multiple stylized keyframes in such cases. However, generating consistent multiple keyframes itself is a challenge. Our method can also be perceived as orthogonal since the (subset of) frames generated by our method can subsequently be used as keyframe inputs to these models.
\begin{figure*}[t]
    \centering
    \includegraphics[width=\textwidth]{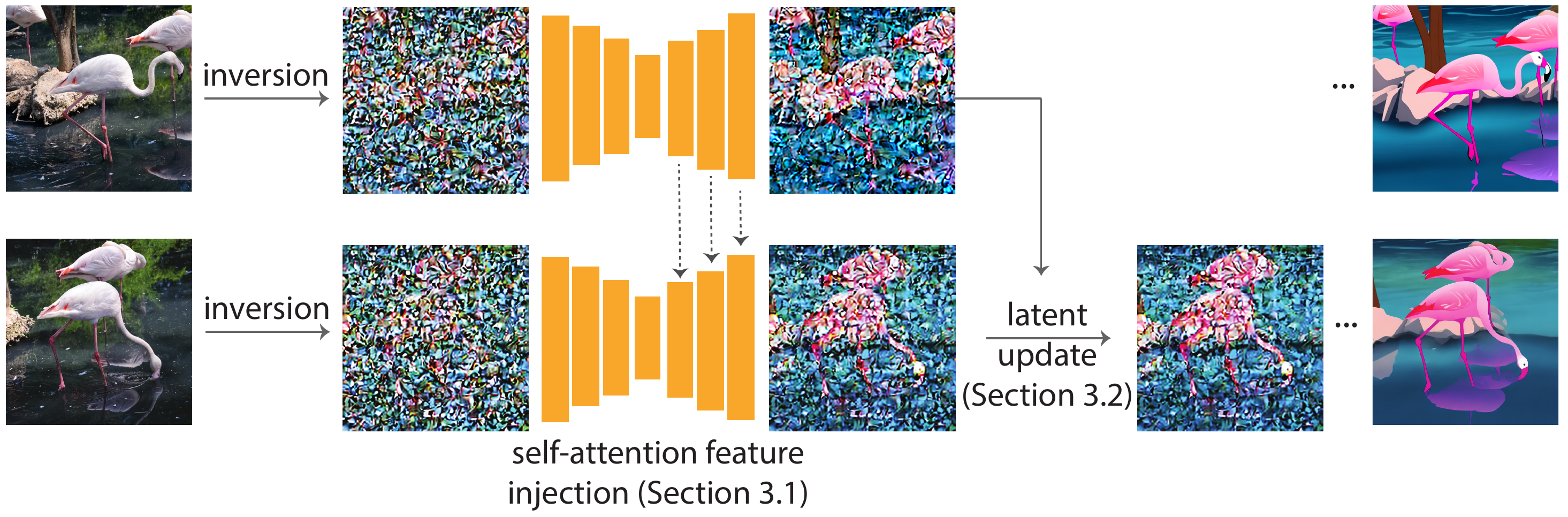}
    \caption{{\bf Method pipeline}. \methodname first inverts each frame with DDIM-inversion and consider it as the initial noise $x_T$ for the denoising process. To edit each frame $i>1$ (lower row), we select a reference frame (upper row), inject its self-attention features to the UNet. At each diffusion step, we also update the latent of the current frame guided by the latent of the reference frame. In practice, we consider both $i-1$ (previous) and $i=1$ (anchor) frames as reference for feature injection, while we use only the previous frame for the guided latent update.}
    \label{fig:pipeline}
\end{figure*}

\section{Our Method}
Given a sequence of frames of a video clip, $\mathcal{I}:=\{I_1,\dots,I_n\}$, we would like to generate a new set of images $\mathcal{I'}:=\{I'_1,\dots,I'_n\}$ that reflects an edit denoted by a target text prompt $\mathcal{P'}$. For example, given a video of a car, the user may want to generate an edited video where attributes of the car, such as its color, are edited. We aim to exploit the power of a pretrained and fixed large-scale image diffusion model to perform such manipulations as coherently as possible, without the need for any example-specific finetuning or extensive training. We achieve this goal by manipulating the internal features of the diffusion model (Section~\ref{sec:attn}) along with additional guidance constraints (Section~\ref{sec:cfg}). 

Given that the fixed image generation model is trained with only single images, it cannot reason about dynamics and geometric changes that happen in an input video. In light of the recent progress in conditioning image generation models with various structure cues~\cite{ediffi,makeascene,voynov2022sketch}, we observe that this additional structure channel is effective in capturing the motion dynamics. Hence, we build our method on a depth-conditioned Stable Diffusion model \cite{SDv2}.
% \footnote{https://huggingface.co/stabilityai/stable-diffusion-2-depth}. 
Given $\mathcal{I}$, we perform per-frame depth prediction~\cite{Ranftl2022} and utilize it as additional input to the model.

%We note that the current state of the art image diffusion models operate either at a low resolution version of the image followed by an upsampling network (in case of pixel diffusion models~\cite{}) or work in the latent space accompanied with an encoder and decoder to map back and forth to the latent space (in case of latent diffusion models~\cite{}). In the rest of the paper, we refer to either these downsampled or latent images via $I_i$ which our method also operates on.

\subsection{Self-attention Feature Injection}
\label{sec:attn}
In the context of static images, a
large-scale image generation diffusion model typically consists of a U-Net architecture composed of residual, self-attention, and cross-attention blocks. While the cross-attention blocks are effective in terms of achieving faithfulness to the text prompt, self-attention layers are effective in determining the overall structure and the appearance of the image. At each diffusion step $t$, the input features $f_t^l$ to the self-attention module at layer $l$, are projected into \textit{queries}, \textit{keys}, and \textit{values} by matrices $W^Q$, $W^K$, and $W^V$, respectively to obtain queries $Q^l$, keys $K^l$, and values $V^l$. The output of the attention block is then computed as:
\begin{align*}
Q^l &= W^Q f_t^l; K^l = W^K f_t^l; V^l = W^v f_t^l \\
\hat{f}_t^l &= \text{softmax} (Q^l (K^l)^T) (V^l).
\end{align*}

In other words, for each location in the current spatial feature map $f_t^l$, a weighted summation of every other spatial features is computed to capture global information. Extending to the context of videos, 
our method captures the interaction across the input image sequence by manipulating the input features to the self-attention module. 
Specifically, we inject the features obtained from the previous frames. A straightforward approach is to attend to the features $f^{j,l}_t$ of an earlier frame $j$ while generating the features $f^{i,l}_t$ for frame $i$ as,
\begin{align*}
Q^{i,l} &= W^Q f_t^{i,l}; K^{i,l} = W^K f_t^{j,l}; V^{i,l} = W^v f_t^{j,l}. \\
\end{align*}
With such feature injection, the current frame is able to utilize the context of the previous frames and hence preserve the appearance changes. A natural question is whether an explicit, potentially recurrent, module can be employed to fuse and represent the state of the previous frame features without explicitly attending to a specific frame. However, the design and training of such a module is not trivial. Instead, we rely on the pre-trained image generation model to perform such fusion implicitly. For each frame $i$, we inject the features obtained from frame $i-1$. 
% Since, during the generation of $i-1$, we attend to frame $i-2$ and so forth, we have an implicit way of aggregating the feature states. 
Since the editing is performed in a frame-by-frame manner, the features of $i-1$ are computed by attending to frame $i-2$. Consequently, we have an implicit way of aggregating the feature states. 
In Section~\ref{sec:results}, we demonstrate that while attending to the previous frame helps to preserve the appearance, in longer sequences it shows the limitation of diminishing the edit. 
%
%The \hl{supplementary material provides} detailed ablations on which previous $j$ frame(s) to use.  Similar to the concurrent work of Wu et al.~\cite{}, we conclude that attending to an \textit{anchor frame} $a$ along with the previous frame $i-1$ results in the most consistent results. Specifically, since $i$ and $i-1$ are more similar, they are a natural choice. However, attending only to the previous frame $i-1$ diminishes the intended edit over longer sequences. 
Attending to an additional anchor frame avoids this forgetful behavior by providing a global constraint on the appearance. Hence, in each self-attention block, we concatenate features from frames $a$ and $i-1$ to compute the key and value pairs. In our experiments, we set $a=1$, i.e., the first frame. %(\duygu{potentially add a figure here} \duygu{experiment with choosing another frame as anchor}):
\begin{align*}
Q^{i,l} &= W^Q f_t^{i,l}; \\
K^{i,l} &= W^K [f_t^{a,l}, f_t^{i-1,l}]; V^{i,l} = W^v [f_t^{a,l}, f_t^{i-1,l}]. \\
\end{align*}

We perform the above feature injection in the decoder layers of the UNet, which we find effective in maintaining appearance consistency. As shown in the ablation study, and also reported by the concurrent work of Tumanyan et al.~\cite{pnpDiffusion2022}, the deeper layers of the decoder capture high resolution and appearance-related information and already result in generated frames with similar appearance but small structural changes. Performing the feature injection in earlier layers of the decoder enables us to avoid such high-frequency structural changes. We do not observe further significant benefit when injecting features in the encoder of the UNet and observe slight artifacts in some examples. %(see the supplementary material). %\duygu{add a figure}. \duygu{which layers do we use?}

\begin{algorithm}[t!]
\hspace*{\algorithmicindent} \textbf{Input} $\mathcal{I} = \{I_1, \dots I_n\}$, text prompt $\mathcal{P'}$, $T=50$ diffusion steps, a pretrained diffusion model SD\\
 \hspace*{\algorithmicindent} \textbf{Output} $\mathcal{I'} = \{I'_1, \dots I'_n\}$
\begin{algorithmic}[1]
\STATE {$\mathcal{X^T} \leftarrow \{ x_T^1, ..., x_T^n\}$ by DDIM inversion }
\STATE {$\mathcal{I'} = \emptyset, \mathcal{F}^{anchor} = \emptyset, \mathcal{F}^{prev} = \emptyset, \mathcal{\hat{X}}_0^{prev} = \emptyset $ }
\FOR{f $\in$ [1,$n$]:}
\STATE{$\mathcal{\hat{X}}_0^{tmp} = [], \mathcal{\hat{F}}^{tmp} = []$}
\STATE{$x_t = x_T^f$}
\STATE{$\delta_{t-1} = 100$ if $t-1 < 25$ else $\delta_{t-1} = 0$}
\FOR{t $\in$ [1,$T$]:}
\IF{$f = 1$}
\STATE{$f^{anchor} = \emptyset, f^{prev} = \emptyset, \hat{x}_0^{prev} = \emptyset$}
\ELSE
\STATE{$f^{anchor} = \mathcal{F}^{anchor}[t]$}
\STATE{$f^{prev} = \mathcal{F}^{prev}[t]$}
\STATE{$\hat{x}_0^{prev} = \mathcal{\hat{X}}_0^{prev}[t]$}
\ENDIF
\STATE{$x_{t-1}, \hat{x}_0^t, f^t = SD(x_t, t, \mathcal{P}, f^{anchor}, f^{prev}) $}
\IF{$f = 1$}
\STATE{$\mathcal{F}^{anchor} \leftarrow  \mathcal{F}^{anchor} \cup \{f^t\}$}
\ELSE
\STATE{$x_{t-1} \leftarrow x_{t-1} - \delta_{t-1} \nabla_{x_t}\| \hat{x}_0^{prev}-\hat{x}_0^t \|^2_2$}
\ENDIF
\STATE{$\mathcal{F}^{tmp} \leftarrow \mathcal{F}^{tmp} \cup \{f^t\}$}

\STATE  {$\mathcal{\hat{X}}_0^{tmp} \leftarrow \mathcal{\hat{X}}_0^{tmp} \cup \{\hat{x}_0^t \} $}
\ENDFOR
\STATE{$\mathcal{\hat{X}}_0^{prev} = \mathcal{\hat{X}}_0^{tmp} $}
\STATE{$\mathcal{F}^{prev} = \mathcal{F}^{tmp} $}
\STATE  {$\mathcal{I'} \leftarrow \mathcal{I'} \cup \{I'_f\} $}
\ENDFOR
\end{algorithmic}
\caption{}
\label{alg}
\end{algorithm}

\subsection{Guided Latent Update}
\label{sec:cfg}
While self-attention feature injection effectively generates frames that have coherent appearance, it can still suffer from temporal flickering. In order to improve the temporal stability, we exploit additional guidance to update the latent variable at each diffusion step along the lines of classifier guidance~\cite{nichol2022glide}. To perform such an update, we first formulate an energy function that enforces consistency.

Stable Diffusion \cite{SDv2,he2022lvdm}, like many other large-scale image diffusion models, is a denoising diffusion implicit model~(DDIM) where at each diffusion step, given a noisy sample $x_t$, a prediction of the noise-free sample $\hat{x}_0$, along with a direction that points to $x_{t}$, is computed. Formally, the final prediction of $x_{t-1}$ is obtained by:
\begin{align*}
x_{t-1} &= \sqrt{\alpha_{t-1}} \underbrace{\hat{x}^t_0}_{\text{predicted `$x_0$'}} + \\
&\underbrace{\sqrt{1-\alpha_{t-1} - \sigma_t^2} \epsilon_{\theta}(x_t, t)}_{\text{direction pointing to $x_t$}} + \underbrace{\sigma_t \epsilon_t}_{\text{random noise}},  \\
\hat{x}^t_0 &= \frac{x_t - \sqrt{1-\alpha_t}\epsilon_{\theta}^t(x_t)}{\sqrt{\alpha_t}},
\end{align*}
where $\alpha_t$ and $\sigma_t$ are the parameters of the scheduler and $\epsilon_{\theta}$ is the noise predicted by the UNet at the current step $t$. %The diffusion process can be made deterministic by setting $\sigma_t=0$. 
The estimate $\hat{x}_0^t$ is computed as a function of $x_t$ and indicates the final generated image. Since our goal is to generate similar consecutive frames eventually, we define an L2 loss function $g(\hat{x}_0^{i,t}, \hat{x}_0^{i-1,t}) = \|\hat{x}_0^{i,t}-\hat{x}_0^{i-1,t}\|^2_2$ that compares the predicted clean images at each diffusion step $t$ between frames $i-1$ and $i$. We update $x_{t-1}^i$, the current noise sample of a frame $i$ at diffusion step $t$, along the direction that minimizes $g$:
\begin{equation}
x_{t-1}^i \leftarrow x_{t-1}^i - \delta_{t-1} \nabla_{x_{t}^i} g(\hat{x}_0^{t,i-1},\hat{x}_0^{t,i}),
\end{equation}
where $\delta_{t-1}$ is a scalar that determines the step size of the update. We empirically set $\delta_{t-1}=100$ in our experiments. We perform this update process for the early denoising steps, namely the first $25$ steps among the total number of $50$ steps, as the overall structure of the generated image is already determined in the earlier diffusion steps \cite{kwon2023semantic}. Performing the latent update in the remaining steps often results in lower-quality images. %, as also reported by others~\cite{xx}.

Finally, the initial noise used to edit each frame also significantly affects the temporal coherency of the generated results. We use an inversion mechanism, DDIM inversion~\cite{song2021ddim}, while other inversion methods aiming to preserve the editability of an image can be used~\cite{mokady2022null} as well. To get a source prompt for inversion, we generate a caption for the first frame of the video using a captioning model~\cite{li2022blip}. We provide the overall steps of our method in Algorithm~\ref{alg}.

\section{Evaluation}
\label{sec:results}
\begin{figure}[t!]
    \centering
    \includegraphics[width=\columnwidth]{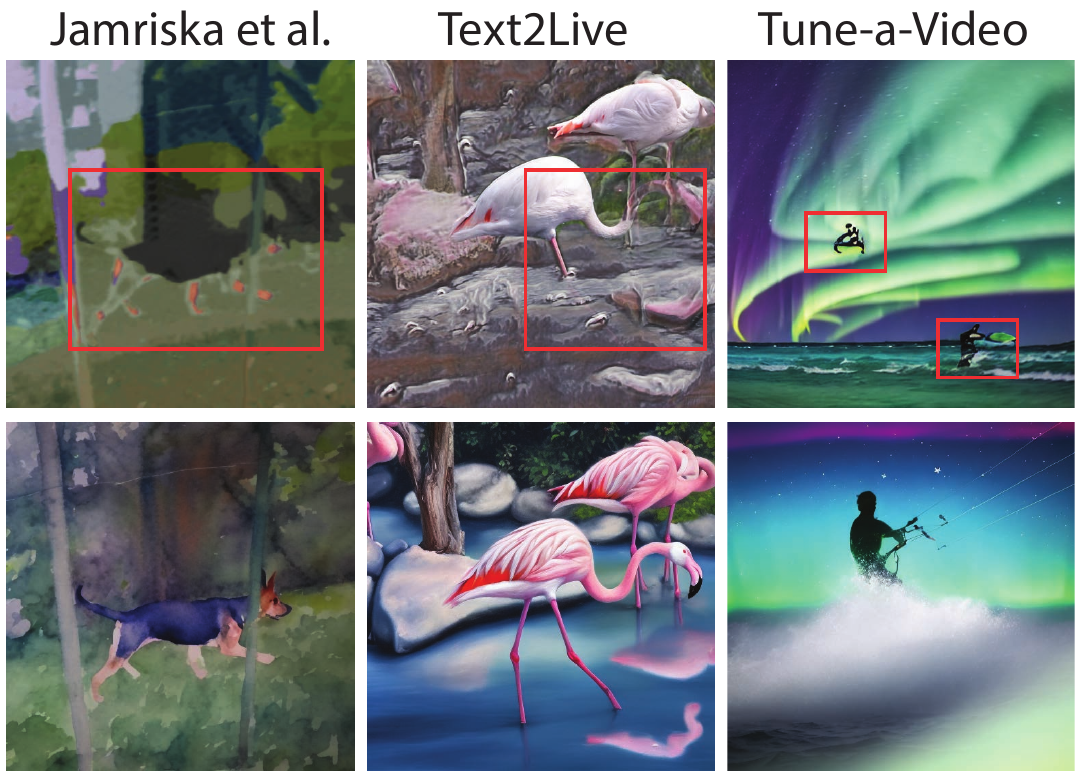}
    \caption{\textbf{Comparisons}. Top: results from baselines; bottom: our results. The method of Jamriska et al.~relies on optical flow to propagate edits in a temporally coherent manner but fails as new content becomes visible. Text2Live suffers when multiple foreground objects are present and a neural atlas cannot be computed robustly. Without an explicit notion of structure, Tune-a-Video is not able to preserve the structure in the input video for some edits.}
    \label{fig:comparison}
\end{figure}

\begin{figure*}
    \centering
   % \begin{overpic}[grid,width=\textwidth]{figures/results_opt.pdf}
%\put(5,95){input video}
%\put(2,85){\texttt{test font}}
%    \end{overpic}
\includegraphics[width=\textwidth]{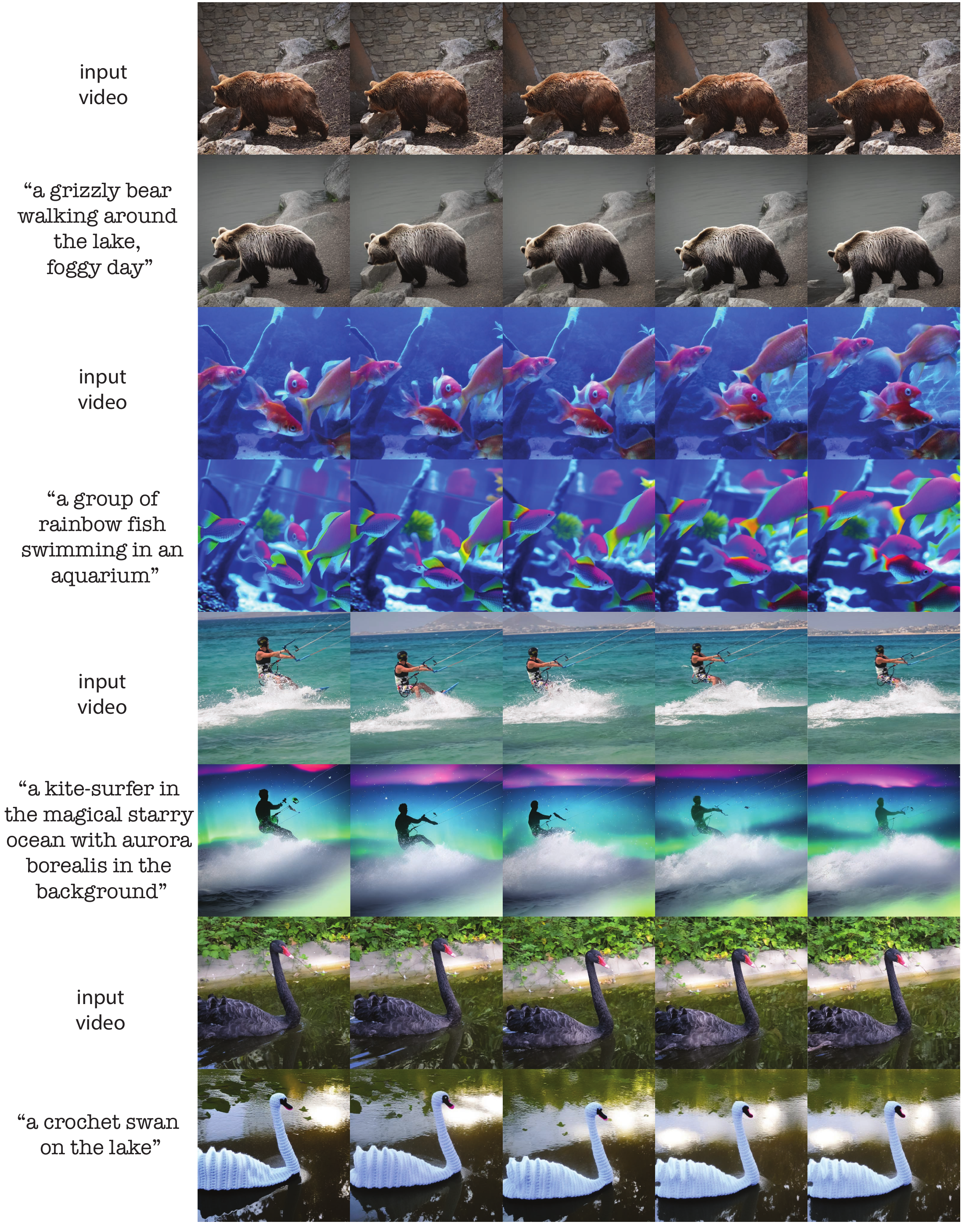}
\vspace*{-.3in}
    \caption{
   \textbf{ Text-guided video edits. }
    For each example, we show frames from the input video at the top and the corresponding edited frames, and the edit prompt at the bottom. Please refer to the supplemental for the videos. }
    \label{fig:results}
\end{figure*}

\textbf{Dataset.} 
Following \cite{bar2022text2live,esser2023structure,wu2022tuneavideo}, we evaluate \methodname on videos obtained from the DAVIS dataset~\cite{Perazzi2016}. %We generate a source prompt for each video by running the BLIP~\cite{li2022blip} captioning model on the first frame. The source prompt is used during the inversion step. 
For videos that have been used in previous or concurrent work, we use editing prompts provided by such work. For other videos, we generate editing prompts by consulting a few users. 
The length of these videos ranges from 50 to 82 frames.

\begin{table}[b!]
\centering
  \caption{Compared to other video-based baselines, our method does not require a heavy pre-processing stage nor a per-video or per-edit finetuning strategy.}
\small
\begin{tabular}{p{2.5cm}||p{1.5cm}|p{1.3cm}|p{1.1cm}  }
 \hline
 & pre-processing & finetuning & layering \\
\hline
 Jamriska et al.~\cite{Jamriska19-SIG}   & no & no &  no\\
Text2Live \cite{bar2022text2live} &   ~7-8 hours  & ~30 min   & yes\\
Tune-a-Video \cite{wu2022tuneavideo} & no & ~10 min &  no\\
ours    & no & no &  no\\
 \hline
 \end{tabular}
 % \caption{We compare our method to several video-based baselines of which some require a heavy pre-processing stage and a per-video or per-edit finetuning strategy.}
 \label{tab:baselines}
\end{table}

\textbf{Baselines.} We compare \methodname with both state-of-the-art image and video editing approaches. 
(i)~The method of Jamriska et al.~\cite{Jamriska19-SIG} propagates the style of a set of given frames to the input video clip. We use the edited anchor frame as a keyframe. 
% Next, we compare to the recent Text2Live~\cite{bar2022text2live} method which provides a method for text-guided editing of neural atlases computed for the foreground and background of a given video clip. 
(ii)~We compare to a recent text-guided video-editing method, Text2Live~\cite{bar2022text2live}. We note that this method first requires the computation of a neural atlas~\cite{kasten2021layered} for the foreground and background layers of a video which takes approximately 7-8 hours per video. Given the neural atlas, the method  further finetunes the text-to-image generation model which takes another 30 minutes.
(iii)~We also compare against SDEdit~\cite{meng2021sdedit} where we add noise to each input frame and denoise conditioned on the edit prompt. We experiment with different strengths of noise added and use the depth-conditioned Stable Diffusion \cite{SDv2} 
% \footnote{https://huggingface.co/stabilityai/stable-diffusion-2-depth} 
as in our backbone diffusion model. 
(iv)~Finally, we also consider the concurrent Tune-a-Video~\cite{wu2022tuneavideo} method, which performs a video-specific finetuning of a pretrained image model. Since this method generates only a limited number of frames, we generate 24 frames by sampling every other frame in the input video following the setup provided by the authors. Note that this method is not conditioned on any structure cues like depth. 
We summarize the characteristics of each baseline in Table~\ref{tab:baselines}.

\textbf{Metrics.} We expect a successful video edit to faithfully reflect the edited prompt and be temporally coherent. To capture the \textit{faithfulness}, we follow \cite{imagen-video,wu2022tuneavideo} and report CLIP score \cite{hessel2021clipscore,park2021benchmark}, which is the cosine similarity between the CLIP embedding \cite{radford2021clip} of the edit prompt and the embedding of each frame in the edited video. 
We refer to this metric as ``CLIP-Text". To measure the \textit{temporal coherency}, we measure the average CLIP similarity between the image embeddings of consecutive frames in the edited video (``CLIP-Image"). We also compute the optical flow between consecutive frames~\cite{RAFT}, and warp each frame in the edited video to the next using the flow. We compute the average mean-squared pixel error between each warped frame and its corresponding target frame as ``Pixel-MSE".

\subsection{Results}
\textbf{Qualitative results.} We provide a set of example edits our method achieves in Figure~\ref{fig:results}. For each example, we show several randomly sampled frames both from the input and edited video, along with the edit prompts. As seen in the figure, \methodname can handle videos with a clear foreground object (e.g., bear) as well as multiple foreground objects (e.g., fish). We can perform \textit{localized} edits where a prompt specifies the attribute of an object (e.g., swan) as well as \textit{global} edits which change the overall style (e.g., kite surfer). Please note that, unlike Text2Live, we do \textit{not} use any explicit mask information to specify which regions should be edited. This enables us to handle reflections automatically, as in the swan example. We refer to the supplementary material for more examples.

\textbf{Quantitative results and comparisons.} We provide quantitative comparisons to the baseline methods in Table~\ref{tab:quanti_result} and also refer to Figure~\ref{fig:comparison}. Among the baseline methods, we observe that Jamriska et al.~\cite{Jamriska19-SIG} achieve good temporal coherency as it explicitly utilizes flow information. However, as new content appears in the video or when flow information is unreliable, it fails to hallucinate details resulting in less faithful edits (Fig.~\ref{fig:comparison} top left). 
Tex2Live~\cite{bar2022text2live} performs well when a clear foreground and background separation exists, and an accurate neural atlas can be synthesized. Since this method edits the atlas itself it is temporally coherent by construction. 
However, when there are multiple foreground objects, e.g.,~Fig.~\ref{fig:comparison} top middle, a reliable neural atlas cannot be computed, and the method fails. The edited results have ``ghost shadow" that consequently deteriorates CLIP-Text scores. 
% \duygu{refer to the faithfullness and pixel metrics}. 
We also observe that, unlike our method, it is not straightforward to perform global style edits on \emph{both} the foreground and background consistently with Text2Live. 
While attaining high CLIP-Text scores, i.e., generating frames faithful to the edit prompt, SDEdit~\cite{meng2021sdedit} results in worst temporal coherency as it generates each frame independently. This is confirmed by the lowest CLIP-Image score and highest Pixel-MSE in Table~\ref{tab:quanti_result}.
% \duygu{talk about the numbers}  
The concurrent Tune-a-Video method~\cite{wu2022tuneavideo} achieves a nice balance between edit propagation and temporal coherency. However, sub-sampling a fixed number of frames in the video inevitably hurts the temporal scores. 
We also observe that for some edits it cannot preserve the structure of the objects in the input video (Fig.~\ref{fig:comparison} top right). Some edits result in very similar outputs, which could be attributed to per-example finetuning that might cause overfitting (see the pig and fish examples in the supplementary material). 
In contrast, by using the additional depth conditioning, \methodname better preserves the structure of the input video and strikes a good balance between respecting the edit as well as keeping temporal consistency without requiring any training. 
% For more discussion and analysis on the comparison, please see supplementary material

\begin{table}
 \caption{
 \textbf{Quantitative comparison.} Our method attains the highest CLIP-Text score (faithfulness) and fairly good CLIP-Image and Pixel-MSE (temporal coherency).}
\footnotesize
\begin{tabular}{lrrr}
\toprule
 & CLIP-Text $\uparrow$ & CLIP-Image $\uparrow$ & Pixel-MSE $\downarrow$ \\
\midrule  \midrule
Jamriska et al.~\cite{Jamriska19-SIG}   & 0.2684 & 0.9838 & 44.62\\
Tex2Live~\cite{bar2022text2live} &   0.2679  &  0.9806  & 72.57\\
Tune-a-Video~\cite{wu2022tuneavideo} & 0.2691 & 0.9674 &  1190.62 \\
SDEdit~\cite{meng2021sdedit}   & 0.2775 & 0.8731 &  2324.29\\
ours w/o update   & 0.2893 & 0.9740 &  371.18\\
ours    & 0.2891 & 0.9767 &  228.62\\
 \bottomrule
 \end{tabular}
 \label{tab:quanti_result}
\end{table}

\textbf{User study.} We further evaluate \methodname against the baselines with a user study. Given 10 videos with 2 different edits each, we ask \totaluser participants to compare our result with one of the baselines shown in random order. Each edited video is ranked, by pairwise comparison, by \eachuser users on average. We ask two questions to the user: (i)~Which video better represents the provided editing caption? (ii)~Which edited video do you prefer? The first question evaluates ``faithfulness" while the second indicates overall video quality via ``preference".
Please see supp.~mat.~for more details on our perceptual study.

\begin{figure}[t]
     \centering
     \begin{subfigure}[t]{0.49\columnwidth}
         \centering
         \includegraphics[trim={1.5cm 0.5cm 1cm 0.1cm},clip,width=\linewidth]{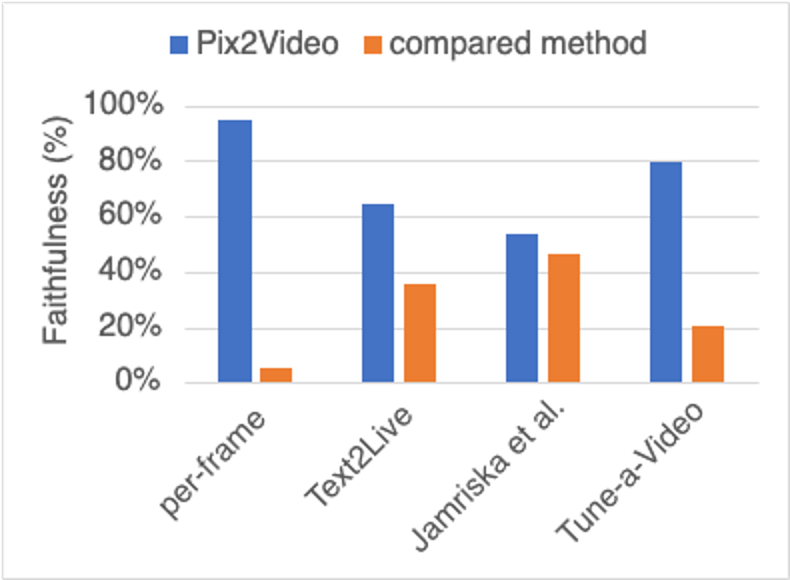}
         \caption{Faithfulness}
         \label{fig:user_faithfulness}
     \end{subfigure}
     \hfill
     \begin{subfigure}[t]{0.49\columnwidth}
         \centering
         \includegraphics[trim={1.5cm 0.5cm 1cm 0.1cm},clip,width=\linewidth]{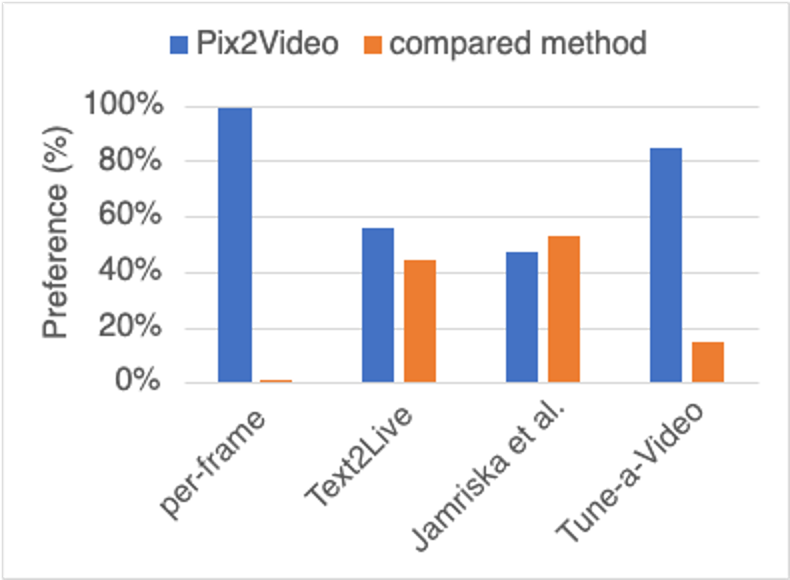}
         \caption{Preference}
         \label{fig:user_preference}
    \end{subfigure}
    % \hfill
    % \begin{subfigure}[t]{\columnwidth}
    %      \centering
    %      \includegraphics[trim={0 0 0 -0.5cm},clip,width=\linewidth]{figures/user_overall.png}
    %      \caption{Chosen frequency (\%) := obtained votes / total occurrence}
    %      \label{fig:user_ratio}
    %  \end{subfigure}     
     \caption{
     \textbf{User evaluation. }
     Our user study shows that \methodname not only better reflects the edits but is also perceptually preferred over other methods. }
\end{figure}

In Fig.~\ref{fig:user_faithfulness}, the majority of the participants agreed that our results reflect the edits more faithfully than others, in accordance with the higher CLIP-Text score in Table~\ref{tab:quanti_result}. 
Fig.~\ref{fig:user_preference} shows that our results are also preferred over other baselines when viewed side-by-side. 
Note that temporal smoothness plays a crucial role in the perceptual quality of a video. Despite Jamriska et al.~\cite{Jamriska19-SIG} losing edits as pointed out in Fig.~\ref{fig:comparison}, it is on par with our method in terms of overall preference which we attribute to high temporal coherency (see Table~\ref{tab:quanti_result}). 
% Finally, Fig.~\ref{fig:user_ratio} shows how often a method is chosen when presented together with other methods. 
In summary, the user study confirms that we achieve a good balance between ensuring edits and maintaining temporal consistency.
% Please \duygu{discuss the results}

\begin{figure}[b!]
    \centering
    \includegraphics[width=\columnwidth]{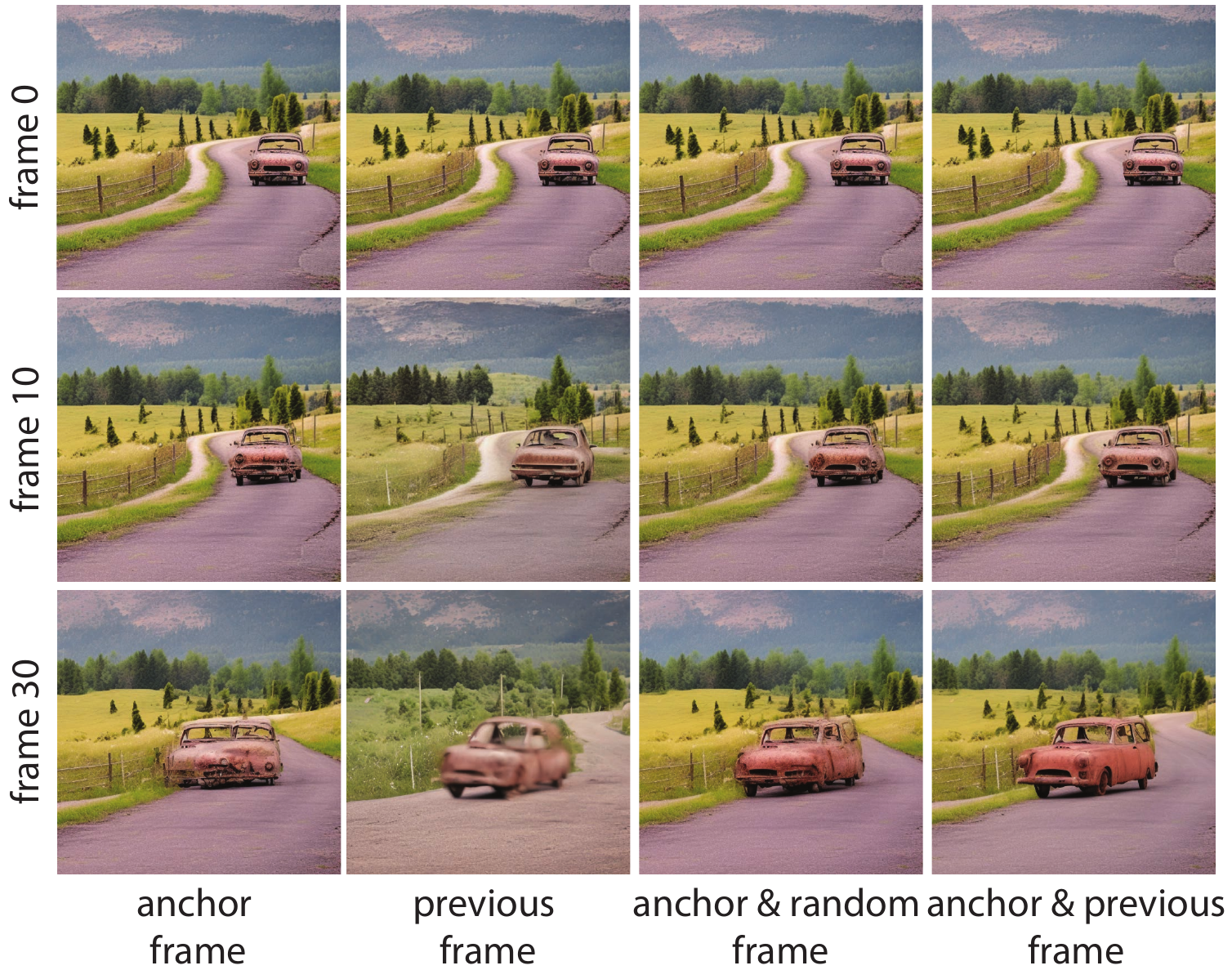}
    \caption{
    \textbf{Ablations. }
    We evaluate different choices for self-attention feature injection. Using a fixed anchor frame results in structural artifacts as the distance between the anchor and the edited frame increases. Attending only to the previous frame or a randomly selected previous frame results in temporal and structural artifacts. We obtain the best results by using    
    a fixed anchor and the previous frame.}
    \label{fig:keyframe}
\end{figure}

\textbf{Ablations.} We perform ablation studies to validate several design choices. First, we evaluate different choices of previous frames to use for self attention feature injection. In Figure \ref{fig:keyframe}, we compare scenarios where we always attend to (i) a fixed anchor frame (first frame in our experiments), (ii) the previous frame only, (iii) an anchor frame and a randomly selected previous frame, and (iv) an anchor frame and a previous frame as in our method. In cases where no previous frame information is used or a random previous frame is chosen, we observe artifacts, especially for sequences that contain more rotational motion, e.g., structure of the car not being preserved as the car rotates. This confirms our intuition that attending to the previous frame implicitly represents the state of the edit in a recurrent manner. Without an anchor frame, we observe more temporal flickering and the edit diminishes as the video progresses. By combining the previous frame with an anchor frame we strike a good balance.

\methodname consists of two main steps of feature injection and guided latent update. In Table~\ref{tab:quanti_result}, we disable the latent update step. As shown in the metrics, this results in worse CLIP-Image scores and Pixel-MSE errors confirming that this guidance is effective in enforcing temporal coherency and preserving the edit. We refer to the supplementary material for more ablation studies.

\textbf{Implementation details.} We use the publicly available Stable Diffusion depth conditioned model~\cite{SDv2} 
% \footnote{https://huggingface.co/stabilityai/stable-diffusion-2-depth} 
as our image generation model. We obtain depth estimates for the input videos using the Midas depth estimator~\cite{Ranftl2022}. For each video, we perform the inversion once and save it. Each editing operation then amounts to generating each frame given the target prompt. We generate results at $512 \times 512$ resolution. The temporal error Pixel-MSE in Table \ref{tab:quanti_result} is therefore computed on a $512 \times 512$ image domain. Our method does not incur any additional significant cost on the image inference step. In our current implementation, we generate each frame in $5$ seconds using a batch size of 1 on an A100 GPU once the frames are inverted in a one time process. 

\section{Conclusion and Future Work}
We present \methodname, a method that utilizes a pre-trained and fixed text-to-image generation model for editing video clips. We demonstrate the power of our method on various inputs and editing tasks. We provide detailed comparisons to baselines along with an extensive user study. We show that \methodname is on par or superior to baselines while not requiring additional pre-processing or finetuning.

However, our method also has limitations that we would like to address in future work. We believe that there is still room for improvement in terms of temporal coherency. Exploiting other energy terms, e.g., patch-based similarity~\cite{Jamriska19-SIG} and CLIP similarity, during the latent update stage, is a promising direction. As we utilize an anchor frame for feature injection, handling longer videos where the distance from the anchor increases can cause quality degradation. Additional conditioning (e.g., image embedding conditioning~\cite{brooks2022instructpix2pix}) and a smart anchor update mechanism is a potential direction. Finally, given that our method does not require any finetuning, it has the advantage of being applied to parallel efforts that aim to introduce additional control to the image generation model, which we would like to exploit.

% \paragraph{Acknowledgment}
% \paragraph{Disclosure}

% \begin{acks}
% We would like to thank ...
% \end{acks}

{\small
\bibliographystyle{ieee_fullname}
% \typeout{} 
\bibliography{main}

\begin{thebibliography}{10}\itemsep=-1pt

\bibitem{SDv2}
{\em Stable Diffusion v2}, 2022.
\newblock \url{https://huggingface.co/stabilityai/stable-diffusion-2-depth}.

\bibitem{ediffi}
Yogesh Balaji, Seungjun Nah, Xun Huang, Arash Vahdat, Jiaming Song, Karsten
  Kreis, Miika Aittala, Timo Aila, Samuli Laine, Bryan Catanzaro, Tero Karras,
  and Ming-Yu Liu.
\newblock {eDiff-I}: Text-to-image diffusion models with an ensemble of expert
  denoisers.
\newblock {\em arXiv preprint arXiv:2211.01324}, 2022.

\bibitem{bar2022text2live}
Omer Bar-Tal, Dolev Ofri-Amar, Rafail Fridman, Yoni Kasten, and Tali Dekel.
\newblock {Text2Live}: Text-driven layered image and video editing.
\newblock In {\em {European Conference on Computer Vision (ECCV)}}, pages
  707--723. Springer, 2022.

\bibitem{brock2018large}
Andrew Brock, Jeff Donahue, and Karen Simonyan.
\newblock Large scale {GAN} training for high fidelity natural image synthesis.
\newblock In {\em {International Conference on Learning Representations
  (ICLR)}}, 2019.

\bibitem{brooks2022generating}
Tim Brooks, Janne Hellsten, Miika Aittala, Ting-Chun Wang, Timo Aila, Jaakko
  Lehtinen, Ming-Yu Liu, Alexei~A Efros, and Tero Karras.
\newblock Generating long videos of dynamic scenes.
\newblock In {\em {Conference on Neural Information Processing Systems
  (NeurIPS)}}, 2022.

\bibitem{brooks2022instructpix2pix}
Tim Brooks, Aleksander Holynski, and Alexei~A Efros.
\newblock Instructpix2pix: Learning to follow image editing instructions.
\newblock {\em arXiv preprint arXiv:2211.09800}, 2022.

\bibitem{Choi2021ILVR}
Jooyoung Choi, Sungwon Kim, Yonghyun Jeong, Youngjune Gwon, and Sungroh Yoon.
\newblock {ILVR}: Conditioning method for denoising diffusion probabilistic
  models.
\newblock In {\em {International Conference on Computer Vision ({ICCV})}},
  pages 14367--14376, October 2021.

\bibitem{croitoru2022diffusion}
Florinel-Alin Croitoru, Vlad Hondru, Radu~Tudor Ionescu, and Mubarak Shah.
\newblock Diffusion models in vision: A survey.
\newblock {\em arXiv preprint arXiv:2209.04747}, 2022.

\bibitem{NEURIPS2021_49ad23d1}
Prafulla Dhariwal and Alexander Nichol.
\newblock Diffusion models beat gans on image synthesis.
\newblock In M. Ranzato, A. Beygelzimer, Y. Dauphin, P.S. Liang, and J.~Wortman
  Vaughan, editors, {\em {Conference on Neural Information Processing Systems
  (NeurIPS)}}, volume~34, pages 8780--8794, 2021.

\bibitem{esser2023structure}
Patrick Esser, Johnathan Chiu, Parmida Atighehchian, Jonathan Granskog, and
  Anastasis Germanidis.
\newblock Structure and content-guided video synthesis with diffusion models.
\newblock {\em arXiv preprint arXiv:2302.03011}, 2023.

\bibitem{makeascene}
Oran Gafni, Adam Polyak, Oron Ashual, Shelly Sheynin, Devi Parikh, and Yaniv
  Taigman.
\newblock {Make-A-Scene}: Scene-based text-to-image generation with human
  priors.
\newblock {\em arXiv}, 2022.

\bibitem{NIPS2014_5ca3e9b1}
Ian Goodfellow, Jean Pouget-Abadie, Mehdi Mirza, Bing Xu, David Warde-Farley,
  Sherjil Ozair, Aaron Courville, and Yoshua Bengio.
\newblock Generative adversarial nets.
\newblock In {\em {Conference on Neural Information Processing Systems
  (NeurIPS)}}, volume~27, 2014.

\bibitem{Gupta_2022_CVPR}
Sonam Gupta, Arti Keshari, and Sukhendu Das.
\newblock {RV-GAN}: Recurrent gan for unconditional video generation.
\newblock In {\em Proceedings of the IEEE/CVF Conference on Computer Vision and
  Pattern Recognition (CVPR) Workshops}, pages 2024--2033, June 2022.

\bibitem{He_2017_ICCV}
Kaiming He, Georgia Gkioxari, Piotr Dollar, and Ross Girshick.
\newblock Mask r-cnn.
\newblock In {\em {International Conference on Computer Vision ({ICCV})}}, Oct
  2017.

\bibitem{he2022lvdm}
Yingqing He, Tianyu Yang, Yong Zhang, Ying Shan, and Qifeng Chen.
\newblock Latent video diffusion models for high-fidelity video generation with
  arbitrary lengths.
\newblock {\em arXiv preprint arXiv:2211.13221}, 2022.

\bibitem{hertz2022prompt}
Amir Hertz, Ron Mokady, Jay Tenenbaum, Kfir Aberman, Yael Pritch, and Daniel
  Cohen-Or.
\newblock Prompt-to-prompt image editing with cross attention control.
\newblock In {\em {International Conference on Learning Representations
  (ICLR)}}, 2023.

\bibitem{hessel2021clipscore}
Jack Hessel, Ari Holtzman, Maxwell Forbes, Ronan~Le Bras, and Yejin Choi.
\newblock Clipscore: A reference-free evaluation metric for image captioning.
\newblock {\em arXiv preprint arXiv:2104.08718}, 2021.

\bibitem{imagen-video}
Jonathan Ho, William Chan, Chitwan Saharia, Jay Whang, Ruiqi Gao, Alexey
  Gritsenko, Diederik~P. Kingma, Ben Poole, Mohammad Norouzi, David~J. Fleet,
  and Tim Salimans.
\newblock Imagen video: High definition video generation with diffusion models.
\newblock {\em arXiv}, 2022.

\bibitem{ho2020ddpm}
Jonathan Ho, Ajay Jain, and Pieter Abbeel.
\newblock Denoising diffusion probabilistic models.
\newblock In {\em {Conference on Neural Information Processing Systems
  (NeurIPS)}}, volume~33, pages 6840--6851, 2020.

\bibitem{hong2022cogvideo}
Wenyi Hong, Ming Ding, Wendi Zheng, Xinghan Liu, and Jie Tang.
\newblock {CogVideo}: Large-scale pretraining for text-to-video generation via
  transformers.
\newblock {\em arXiv preprint arXiv:2205.15868}, 2022.

\bibitem{Jamriska19-SIG}
Ond\v{r}ej Jamri\v{s}ka, \v{S}\'{a}rka Sochorov\'{a}, Ond\v{r}ej Texler, Michal
  Luk\'{a}\v{c}, Jakub Fi\v{s}er, Jingwan Lu, Eli Shechtman, and Daniel
  S\'{y}kora.
\newblock Stylizing video by example.
\newblock {\em ACM Transactions on Graphics}, 38(4), 2019.

\bibitem{Karras2022edm}
Tero Karras, Miika Aittala, Timo Aila, and Samuli Laine.
\newblock Elucidating the design space of diffusion-based generative models.
\newblock In {\em {Conference on Neural Information Processing Systems
  (NeurIPS)}}, 2022.

\bibitem{stylegan}
Tero Karras, Samuli Laine, and Timo Aila.
\newblock A style-based generator architecture for generative adversarial
  networks.
\newblock In {\em {Computer Vision and Pattern Recognition (CVPR)}}, pages
  4401--4410, 2019.

\bibitem{kasten2021layered}
Yoni Kasten, Dolev Ofri, Oliver Wang, and Tali Dekel.
\newblock Layered neural atlases for consistent video editing.
\newblock {\em ACM Transactions on Graphics (TOG)}, 40(6):1--12, 2021.

\bibitem{kawar2022imagic}
Bahjat Kawar, Shiran Zada, Oran Lang, Omer Tov, Huiwen Chang, Tali Dekel, Inbar
  Mosseri, and Michal Irani.
\newblock Imagic: Text-based real image editing with diffusion models.
\newblock {\em arXiv preprint arXiv:2210.09276}, 2022.

\bibitem{kwon2023semantic}
Mingi Kwon, Jaeseok Jeong, and Youngjung Uh.
\newblock Diffusion models already have a semantic latent space.
\newblock In {\em {International Conference on Learning Representations
  (ICLR)}}, 2023.

\bibitem{li2022blip}
Junnan Li, Dongxu Li, Caiming Xiong, and Steven Hoi.
\newblock {BLIP}: Bootstrapping language-image pre-training for unified
  vision-language understanding and generation.
\newblock In {\em ICML}, 2022.

\bibitem{lu2020}
Erika Lu, Forrester Cole, Tali Dekel, Weidi Xie, Andrew Zisserman, David
  Salesin, William~T Freeman, and Michael Rubinstein.
\newblock Layered neural rendering for retiming people in video.
\newblock In {\em SIGGRAPH Asia}, 2020.

\bibitem{meng2021sdedit}
Chenlin Meng, Yutong He, Yang Song, Jiaming Song, Jiajun Wu, Jun-Yan Zhu, and
  Stefano Ermon.
\newblock {SDEdit}: Guided image synthesis and editing with stochastic
  differential equations.
\newblock In {\em {International Conference on Learning Representations
  (ICLR)}}, 2021.

\bibitem{mokady2022null}
Ron Mokady, Amir Hertz, Kfir Aberman, Yael Pritch, and Daniel Cohen-Or.
\newblock Null-text inversion for editing real images using guided diffusion
  models.
\newblock {\em arXiv preprint arXiv:2211.09794}, 2022.

\bibitem{Dreamix}
Eyal Molad, Eliahu Horwitz, Dani Valevski, Alex~Rav Acha, Yossi Matias, Yael
  Pritch, Yaniv Leviathan, and Yedid Hoshen.
\newblock Dreamix: Video diffusion models are general video editors.
\newblock {\em arXiv}, 2023.

\bibitem{t2iadapter}
Chong Mou, Xintao Wang, Liangbin Xie, Jian Zhang, Zhongang Qi, Ying Shan, and
  Xiaohu Qie.
\newblock T2i-adapter: Learning adapters to dig out more controllable ability
  for text-to-image diffusion models.
\newblock {\em arXiv}, 2023.

\bibitem{nichol2022glide}
Alexander~Quinn Nichol, Prafulla Dhariwal, Aditya Ramesh, Pranav Shyam, Pamela
  Mishkin, Bob McGrew, Ilya Sutskever, and Mark Chen.
\newblock {GLIDE:} towards photorealistic image generation and editing with
  text-guided diffusion models.
\newblock In {\em International Conference on Machine Learning, {ICML} 2022,
  17-23 July 2022, Baltimore, Maryland, {USA}}, volume 162, pages 16784--16804,
  2022.

\bibitem{park2021benchmark}
Dong~Huk Park, Samaneh Azadi, Xihui Liu, Trevor Darrell, and Anna Rohrbach.
\newblock Benchmark for compositional text-to-image synthesis.
\newblock In {\em Neural Information Processing Systems Datasets and Benchmarks
  Track}, 2021.

\bibitem{Perazzi2016}
F. Perazzi, J. Pont-Tuset, B. McWilliams, L. {Van Gool}, M. Gross, and A.
  Sorkine-Hornung.
\newblock A benchmark dataset and evaluation methodology for video object
  segmentation.
\newblock In {\em Computer Vision and Pattern Recognition}, 2016.

\bibitem{StyleFaceV}
Haonan Qiu, Yuming Jiang, Hang Zhou, Wayne Wu, and Ziwei Liu.
\newblock Stylefacev: Face video generation via decomposing and recomposing
  pretrained stylegan3.
\newblock {\em arXiv}, 2022.

\bibitem{radford2021clip}
Alec Radford, Jong~Wook Kim, Chris Hallacy, Aditya Ramesh, Gabriel Goh,
  Sandhini Agarwal, Girish Sastry, Amanda Askell, Pamela Mishkin, Jack Clark,
  et~al.
\newblock Learning transferable visual models from natural language
  supervision.
\newblock In {\em {International Conference on Machine Learning (ICML)}}, pages
  8748--8763, 2021.

\bibitem{ramesh2022dalle2}
Aditya Ramesh, Prafulla Dhariwal, Alex Nichol, Casey Chu, and Mark Chen.
\newblock Hierarchical text-conditional image generation with clip latents.
\newblock {\em arXiv preprint arXiv:2204.06125}, 2022.

\bibitem{Ranftl2022}
Ren\'{e} Ranftl, Katrin Lasinger, David Hafner, Konrad Schindler, and Vladlen
  Koltun.
\newblock Towards robust monocular depth estimation: Mixing datasets for
  zero-shot cross-dataset transfer.
\newblock {\em {Transactions on Pattern Analysis and Machine Intelligence
  (TPAMI)}}, 44(3), 2022.

\bibitem{rombach2022stablediffusion}
Robin Rombach, Andreas Blattmann, Dominik Lorenz, Patrick Esser, and Bj{\"o}rn
  Ommer.
\newblock High-resolution image synthesis with latent diffusion models.
\newblock In {\em {Computer Vision and Pattern Recognition (CVPR)}}, pages
  10684--10695, 2022.

\bibitem{RuderDB2016}
Manuel Ruder, Alexey Dosovitskiy, and Thomas Brox.
\newblock Artistic style transfer for videos.
\newblock In {\em German Conference on Pattern Recognition}, pages 26--36,
  2016.

\bibitem{saharia2022palette}
Chitwan Saharia, William Chan, Huiwen Chang, Chris~A. Lee, Jonathan Ho, Tim
  Salimans, David~J. Fleet, and Mohammad Norouzi.
\newblock Palette: Image-to-image diffusion models.
\newblock In {\em {International Conference on Computer Graphics and
  Interactive Techniques (SIGGRAPH)}}, 2022.

\bibitem{saharia2022imagen}
Chitwan Saharia, William Chan, Saurabh Saxena, Lala Li, Jay Whang, Emily
  Denton, Seyed Kamyar~Seyed Ghasemipour, Raphael Gontijo-Lopes, Burcu~Karagol
  Ayan, Tim Salimans, Jonathan Ho, David~J. Fleet, and Mohammad Norouzi.
\newblock Photorealistic text-to-image diffusion models with deep language
  understanding.
\newblock In {\em {Conference on Neural Information Processing Systems
  (NeurIPS)}}, 2022.

\bibitem{TGAN2017}
Masaki Saito, Eiichi Matsumoto, and Shunta Saito.
\newblock Temporal generative adversarial nets with singular value clipping.
\newblock In {\em ICCV}, 2017.

\bibitem{singer2023makeavideo}
Uriel Singer, Adam Polyak, Thomas Hayes, Xi Yin, Jie An, Songyang Zhang, Qiyuan
  Hu, Harry Yang, Oron Ashual, Oran Gafni, Devi Parikh, Sonal Gupta, and Yaniv
  Taigman.
\newblock {Make-A-Video}: Text-to-video generation without text-video data.
\newblock In {\em {International Conference on Learning Representations
  (ICLR)}}, 2023.

\bibitem{stylegan-v}
Ivan Skorokhodov, Sergey Tulyakov, and Mohamed Elhoseiny.
\newblock Stylegan-v: A continuous video generator with the price, image
  quality and perks of stylegan2.
\newblock In {\em {Computer Vision and Pattern Recognition (CVPR)}}, pages
  3626--3636, 2022.

\bibitem{song2021ddim}
Jiaming Song, Chenlin Meng, and Stefano Ermon.
\newblock Denoising diffusion implicit models.
\newblock In {\em {International Conference on Learning Representations
  (ICLR)}}, 2021.

\bibitem{RAFT}
Zachary Teed and Jia Deng.
\newblock {RAFT}: Recurrent all-pairs field transforms for optical flow.
\newblock In {\em {European Conference on Computer Vision (ECCV)}}, 2020.

\bibitem{Texler2020}
Ond\v{r}ej Texler, David Futschik, Michal Ku\v{c}era, Ond\v{r}ej Jamri\v{s}ka,
  \v{S}\'{a}rka Sochorov\'{a}, Meglei Chai, Sergey Tulyakov, and Daniel
  S\'{y}kora.
\newblock Interactive video stylization using few-shot patch-based training.
\newblock {\em ACM Transactions on Graphics}, 39(4):73, 2020.

\bibitem{MoCoGAN}
Sergey Tulyakov, Ming-Yu Liu, Xiaodong Yang, and Jan Kautz.
\newblock {MoCoGAN}: Decomposing motion and content for video generation.
\newblock In {\em {Computer Vision and Pattern Recognition (CVPR)}}, pages
  1526--1535, 2018.

\bibitem{pnpDiffusion2022}
Narek Tumanyan, Michal Geyer, Shai Bagon, and Tali Dekel.
\newblock Plug-and-play diffusion features for text-driven image-to-image
  translation.
\newblock {\em arXiv preprint arXiv:2211.12572}, 2022.

\bibitem{villegas2023phenaki}
Ruben Villegas, Mohammad Babaeizadeh, Pieter-Jan Kindermans, Hernan Moraldo,
  Han Zhang, Mohammad~Taghi Saffar, Santiago Castro, Julius Kunze, and Dumitru
  Erhan.
\newblock Phenaki: Variable length video generation from open domain textual
  descriptions.
\newblock In {\em {International Conference on Learning Representations
  (ICLR)}}, 2023.

\bibitem{voynov2022sketch}
Andrey Voynov, Kfir Abernan, and Daniel Cohen-Or.
\newblock Sketch-guided text-to-image diffusion models.
\newblock {\em arXiv preprint arXiv:2211.13752}, 2022.

\bibitem{wang2022pretraining}
Tengfei Wang, Ting Zhang, Bo Zhang, Hao Ouyang, Dong Chen, Qifeng Chen, and
  Fang Wen.
\newblock Pretraining is all you need for image-to-image translation.
\newblock {\em arXiv}, 2022.

\bibitem{wang2018fewshotvid2vid}
Ting-Chun Wang, Ming-Yu Liu, Andrew Tao, Guilin Liu, Jan Kautz, and Bryan
  Catanzaro.
\newblock Few-shot video-to-video synthesis.
\newblock In {\em {Conference on Neural Information Processing Systems
  (NeurIPS)}}, 2019.

\bibitem{wang2018vid2vid}
Ting-Chun Wang, Ming-Yu Liu, Jun-Yan Zhu, Guilin Liu, Andrew Tao, Jan Kautz,
  and Bryan Catanzaro.
\newblock Video-to-video synthesis.
\newblock In {\em {Conference on Neural Information Processing Systems
  (NeurIPS)}}, 2018.

\bibitem{wu2022tuneavideo}
Jay~Zhangjie Wu, Yixiao Ge, Xintao Wang, Stan~Weixian Lei, Yuchao Gu, Wynne
  Hsu, Ying Shan, Xiaohu Qie, and Mike~Zheng Shou.
\newblock {Tune-A-Video}: One-shot tuning of image diffusion models for
  text-to-video generation.
\newblock {\em arXiv preprint arXiv:2212.11565}, 2022.

\bibitem{xu2022videoeditgan}
Yiran Xu, Badour AlBahar, and Jia-Bin Huang.
\newblock Temporally consistent semantic video editing.
\newblock {\em arXiv preprint arXiv: 2206.10590}, 2022.

\bibitem{MAGVIT}
Lijun Yu, Yong Cheng, Kihyuk Sohn, José Lezama, Han Zhang, Huiwen Chang,
  Alexander~G. Hauptmann, Ming-Hsuan Yang, Yuan Hao, Irfan Essa, and Lu Jiang.
\newblock {MAGVIT}: Masked generative video transformer.
\newblock {\em arXiv}, 2022.

\bibitem{yu2023video}
Sihyun Yu, Kihyuk Sohn, Subin Kim, and Jinwoo Shin.
\newblock Video probabilistic diffusion models in projected latent space.
\newblock In {\em {Computer Vision and Pattern Recognition (CVPR)}}, 2023.

\bibitem{digan}
Sihyun Yu, Jihoon Tack, Sangwoo Mo, Hyunsu Kim, Junho Kim, Jung-Woo Ha, and
  Jinwoo Shin.
\newblock Generating videos with dynamics-aware implicit generative adversarial
  networks.
\newblock In {\em {International Conference on Learning Representations
  (ICLR)}}, 2022.

\bibitem{zhang2023controlnet}
Lvmin Zhang and Maneesh Agrawala.
\newblock Adding conditional control to text-to-image diffusion models.
\newblock {\em arXiv preprint arXiv:2302.05543}, 2023.

\bibitem{zhang2022towards}
Qihang Zhang, Ceyuan Yang, Yujun Shen, Yinghao Xu, and Bolei Zhou.
\newblock Towards smooth video composition.
\newblock {\em International Conference on Learning Representations (ICLR)},
  2023.

\end{thebibliography}
}

\clearpage
\begin{appendices} 
\label{appendices}
In this document, we provide details of the perceptual study, the comparison setup, and provide more ablations. We refer the reader to the accompanying webpage for video based results.

\section{Comparison Details.}
We provide more details about the comparisons we run. When comparing to Text2Live~\cite{bar2022text2live}, we use the neural atlases provided by the authors for the \textit{swan}, \textit{dog}, and \textit{car} examples. For other examples, we first compute a mask that is required for neural atlas computation. For computing the masks, we either use MaskRCNN~\cite{He_2017_ICCV} for categories detected by it or perform per-frame foreground object selection\footnote{Photoshop Select Subject}. When comparing to Tune-a-Video~\cite{wu2022tuneavideo}, we use the implementation provided by the authors and finetune on each example for 500 iterations using the default settings.

\section{Perceptual Study}
\begin{figure}[b]
     \centering
    % \begin{subfigure}[t]{\columnwidth}
         \includegraphics[width=\columnwidth]{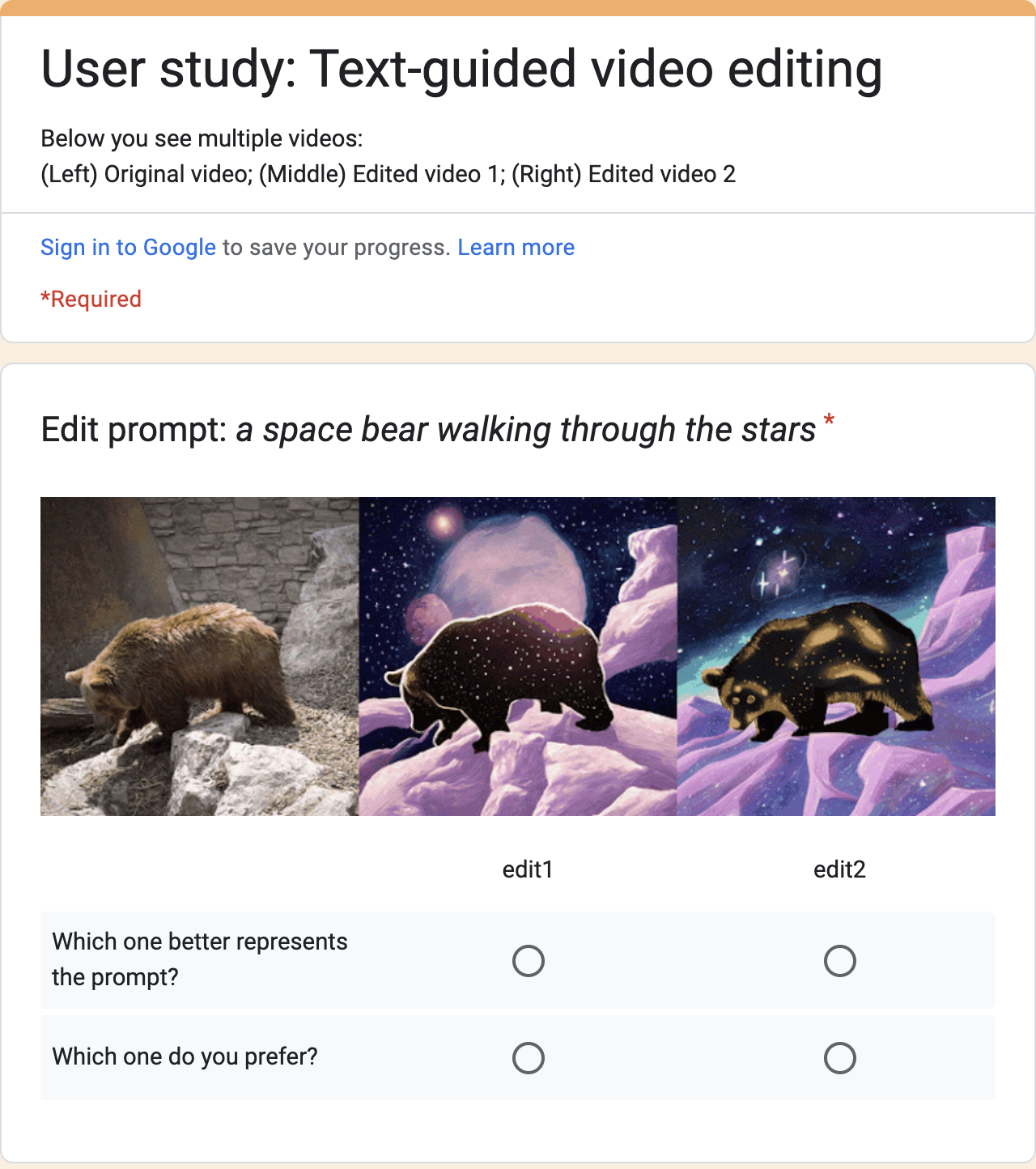}
         % \caption{Question design}
         
     % \end{subfigure}     
     % \begin{subfigure}[t]{\columnwidth}
     %     \centering
     %     \includegraphics[trim={0 0 0 -0.5cm},clip,width=\linewidth]{figures/user_overall.png}
     %     \caption{Chosen frequency (\%) := obtained votes / total occurrence}
     %     \label{fig:user_ratio}
     % \end{subfigure}     
     \caption{
     \textbf{Question design of the user evaluation. }
        }
        \label{fig:user_UI}
\end{figure}

In Figure \ref{fig:user_UI} we show an example of our survey question. 
The input video is on the left; the middle and the right columns are two edited results for comparison, one of which is \methodname while the other one is randomly chosen from 4 baselines: per-frame editing, Jamriska et al.~\cite{Jamriska19-SIG}, 
Text2Live \cite{bar2022text2live} and a concurrent method Tune-a-Video \cite{wu2022tuneavideo}.
Two edited results are placed randomly in the middle or the right to avoid any bias.
We ask two questions: (i)~Which one better represents the prompt (shown on the top)? (ii)~Which one do you prefer?
In the second question we do not ask users to pay particular attention to any attribute, e.g., temporal smoothness or realism, as we aim to evaluate generally the perceptual quality of a video.

% Fig.~\ref{fig:user_ratio} shows how often a method is chosen when presented together with other methods. 

\section{Ablation Study}
\subsection{Effect of layer choice in feature injection.}
\begin{figure*}[t!]
\centering
    \includegraphics[width=\textwidth]{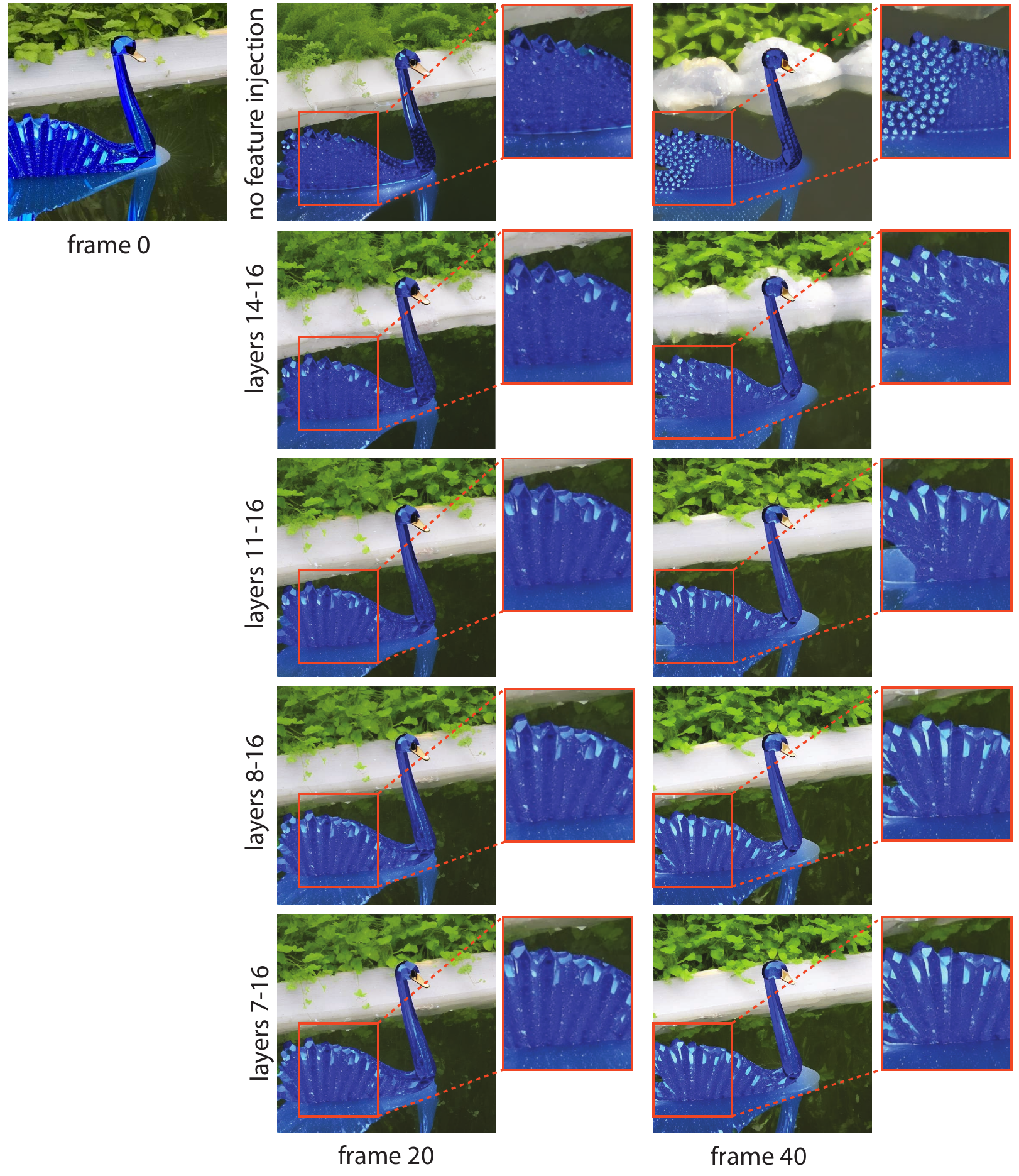}
    \vspace*{-.2in}
    \caption{
     We show the effect of feature injection in different layers of the UNet. Injecting features into the decoder (layers 8-16) help to improve the consistency. We do not observe significant improvements when feature injection is performed for the bottleneck as well (layer 7).
        }
    \label{fig:feature}
\end{figure*}

The Unet of the Stable Diffusion v2 model consists of 16 layers where each layer has a resnet, self-attention, and cross attention modules. 6 of these blocks are part of the encoder, 9 are part of the decoder, and 1 consists of the bottleneck. In Figure~\ref{fig:feature}, we show the effect of applying feature injection into different self attention layers. We observe that injection features into deeper layers of the decoder (i.e., layers 13-16) already results in significant structural and appearance consistency. Injecting features in other decoder layers results in improvements especially in terms of preserving high frequency details. We do not observe further improvements as we inject features into the bottleneck layer. In Table~\ref{tab:quanti_feature}, we compare quantitatively the case where we perform feature injection only in the decoder vs all layers. We observe that while injecting features at all layers results in on-par CLIP-Image and slightly better Pixel-MSE errors, it tends to generate slightly more blurry output. This is also reflected in the slightly worse CLIP-Text errors. Hence, we choose to apply feature injection only for the decoder which strikes a good balance.

\begin{table}
\centering
 \caption{
 We quantitatively compare cased where we perform feature injection at all layers of the UNet vs only the decoder.}
\footnotesize
\begin{tabular}{lrrr}
\toprule
 & CLIP-Text $\uparrow$ & CLIP-Image $\uparrow$ & Pixel-MSE $\downarrow$ \\
\midrule  \midrule
all layers   & 0.2877 & 0.9766 &  200.70\\
decoder    & 0.2891 & 0.9767 &  228.62\\
 \bottomrule
 \end{tabular}
 \label{tab:quanti_feature}
\end{table}

\begin{figure}
\centering
    \includegraphics[width=\columnwidth]{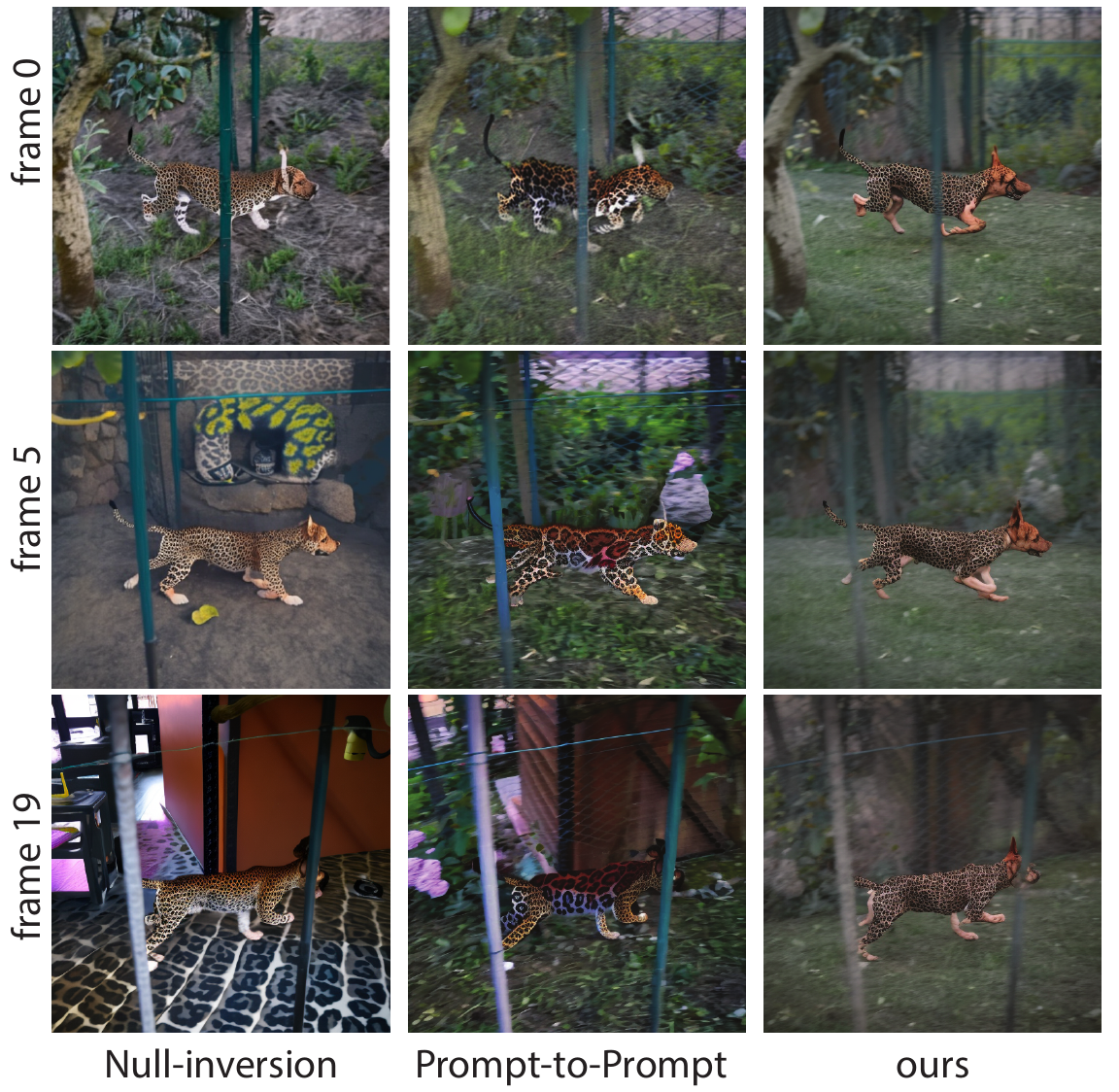}
    \caption{
     We compare our method to additional image based editing methods, Null-inversion~\cite{mokady2022null} and Prompt-to-Prompt~\cite{hertz2022prompt}. When such methods are applied naively in a per-frame manner, they yield inconsistent appearance across frames.
        }
    \label{fig:comp}
\end{figure}

\section{More Results and Discussion}
We compare our method to additional image-based editing methods that are applied per-frame. Specifically, we run the Stable Diffusion depth-to-image pipeline in conjuction with Null-inversion~\cite{mokady2022null} as well as the Prompt-to-Prompt~\cite{hertz2022prompt} editing pipeline as shown in Figure~\ref{fig:comp} and the supplementary webpage. In both cases the per-frame methods result in inconsistent results as expected. While Prompt2Prompt can ensure more localized edits, it cannot guarantee consistency across images.

% Since our method does not require finetuning, it can be adopted with other base generation models. In order to demonstrate this, we implement our method along with Prompt-to-Prompt as shown in Figure~\ref{}. While our method performs feature injection in the self attention layers and a guided latent update, Prompt2Prompt is used for altering the cross attention layers based on the editing prompt. This combination can perform localized edits effectively while ensuring consistency across the images.

Our method uses depth as a structural cue which helps to preserve the structure of the input video. Hence, it can be used in conjunction with style propagation methods as a post processing to further improve the results. Specifically, we provide the every 3rd frame generated by our method as a keyframe to the method of Jamriska et al.~\cite{Jamriska19-SIG} and propagate the style of these keyframes to the inbetween frames. We show that this results in better temporal stability while not suffering from the artifacts that are due to visibility changes when using only a single keyframe (please see the supplementary webpage). We note that, such a postprocessing cannot be applied to Tune-a-video~\cite{wu2022tuneavideo} since the edits do not preserve the original structure and hence using the optical flow obtained from the original video introduces artifacts.

\end{appendices}

\end{document}